\documentclass[sigconf]{acmart}
\AtBeginDocument{%
  }

\setcopyright{acmlicensed}
\copyrightyear{2025}
\acmYear{2025}
\acmDOI{XXXXXXX.XXXXXXX}
\acmConference[Conference acronym 'XX]{Make sure to enter the correct
  conference title from your rights confirmation emai}{June 03--05,
  2025}{Woodstock, NY}
\acmISBN{978-1-4503-XXXX-X/18/06}

 \usepackage{url}
\usepackage{textcomp}
\usepackage{stfloats}
\usepackage{url}
\usepackage{verbatim}
\usepackage{threeparttable}
\usepackage{multirow}
\usepackage{makecell}
\usepackage{tabularx,booktabs}
\usepackage{caption}
\usepackage{amsthm}
\usepackage{bbm}
\usepackage{booktabs}
\usepackage[ruled]{algorithm2e}

\usepackage{subfigure}
\usepackage{multirow}
\usepackage{graphicx}
\usepackage{textcomp}
\usepackage{xcolor}
\usepackage{setspace}
\usepackage[export]{adjustbox}
\usepackage[figuresright]{rotating}
\usepackage{booktabs}
\usepackage{geometry}
\usepackage{arydshln}
\usepackage{pifont}

\usepackage{amsthm}

\usepackage{lineno,hyperref}
\usepackage{multirow}

\begin{document}

\title{Federated Deep Subspace Clustering}

\author{Yupei Zhang}
\email{ypzhaang@nwpu.edu.cn}
\orcid{0000-0001-8348-0545}
\affiliation{%
  \institution{Northwestern Polytechnical University}
  \city{Xi'an}
  \state{Shaanxi}
  \country{China}
}

\author{Ruojia Feng}
\email{2022300396@mail.nwpu.edu.cn}
\authornote{Both authors contributed equally to this research.}

\author{Yifei Wang}
\authornotemark[1]
\email{wang_yf@mail.nwpu.edu.cn}
\affiliation{%
  \institution{Northwestern Polytechnical University}
  \city{Xi'an}
  \state{Shaanxi}
  \country{China}
  }

\author{Xuequn Shang}
\email{shang@mail.nwpu.edu.cn}
\affiliation{%
  \institution{Northwestern Polytechnical University}
  \city{Xi'an}
  \state{Shaanxi}
  \country{China}
  }





\renewcommand{\shortauthors}{Yupei et al.}

\begin{abstract}
This paper introduces FDSC, a private-protected subspace clustering (SC) approach with federated learning (FC) schema. In each client, there is a deep subspace clustering network accounting for grouping the isolated data, composed of a encode network, a self-expressive layer, and a decode network. FDSC is achieved by uploading the encode network to communicate with other clients in the server. Besides, FDSC is also enhanced by preserving the local neighborhood relationship in each client. With the effects of federated learning and locality preservation, the learned data features from the encoder are boosted so as to enhance the self-expressiveness learning and result in better clustering performance. Experiments test FDSC on public datasets and compare with other clustering methods, demonstrating the effectiveness of FDSC.
\end{abstract}




\maketitle

\section{Introduction}

\begin{figure}[t]
\centering
\includegraphics[width=0.5\textwidth]{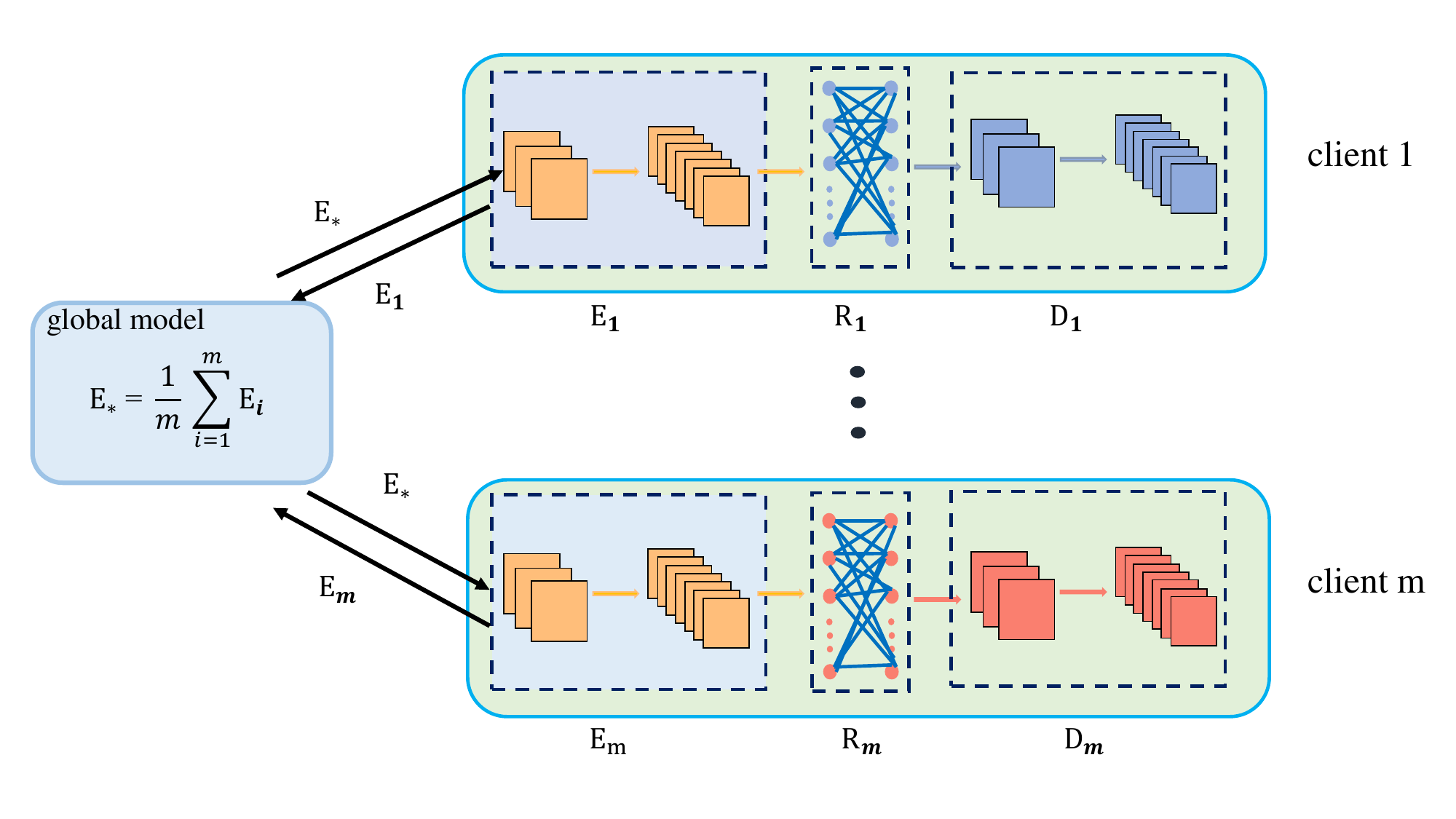}
\caption{FDSC framework. Each client contains shared encoder, private self-expressive layer and private decoder. The server calculates the weighted average for the encoders.}
\label{fig1}
\end{figure}

Subspace clustering (SC) is one of the popular clustering methods in recent years, which aims at learning the potential subspace of high-dimensional data and partitioning the data \cite{peng2020deep}. SC has drawn extensive attention in fields such as spectral clustering \cite{zhang2016spectral}, image segmentation \cite{baek2021deep} and data dimensionality reduction technique \cite{zhang2017low} in the past decades. Recently, a large number of algorithms for spectral clustering have been proposed \cite{li2021self,berahmand2022novel}. Their basic idea is to construct an affinity matrix on the datasets and cluster according to the eigenvectors of the affinity matrix \cite{ng2001spectral}.  

Since real-world data often have complex distribution and non-linear characteristics, spectral clustering in linear space cannot adapt to the actual application scenarios. Deep neural network has the potential to extract data features and map them into linear subspace \cite{li2022neural}, which consists of auto-encoder and self-expressive layer. The auto-encoder exploits self-reconstruction loss to learn latent features \cite{zhu2019multi}, and the self-expressive layer is used to learn the affinity matrix \cite{peng2020deep,lv2021pseudo}. Deep subspace clustering has general application, such as face image segmentation under different lighting and postures \cite{ji2017deep}. Although the performance of deep subspace clustering has been greatly improved through the efforts of all researchers, the rapid increase of data that raises a large number of parameters limits the deep clustering effect \cite{zhang2018survey}. In addition, the decentralization and privacy of real-world data make centralized deep clustering difficult to perform \cite{stallmann2022towards}. The emergence of federated learning is one of the effective ways to solve these problems

Federated learning (FL) is a machine learning framework that enables many clients collaboratively learning a shared model without exchanging their private data \cite{yang2019federated,banabilah2022federated}. Using privacy protection, FL can be applied to most practical application scenarios such as smart healthcare \cite{nguyen2022federated,arikumar2022fl}, movie recommendation \cite{gao2020privacy} and smart retail \cite{banabilah2022federated}. Although FL has been studied a lot in the supervised learning environment, the FL research in unsupervised setting is still little \cite{dennis2021heterogeneity}, especially in the field of clustering.

The main challenge in FL is decentralized and heterogeneous characteristics in the data \cite{han2022fedx}.
In the unsupervised FL setting, clients are usually edge devices \cite{lubana2022orchestra}, which makes data features distributed in different spaces. In order to solve the problem of heterogeneous data, most researches focus on clustering clients, i.e., the clustered federated learning \cite{ghosh2020efficient,sattler2020clustered}. They usually capture the heterogeneity between clients and assign clients to different global models \cite{long2023multi}. However, most methods do not consider the communication problems in FL, resulting in low training efficiency and complicated process \cite{ghosh2020efficient}. A few studies focus on collaborative clustering of data on clients. k-FED \cite{dennis2021heterogeneity} proposed a one-shot federated clustering method based on k-means in which all clients share a set of global cluster centers. However, there is still a lack of research on learning local cluster centers. Generally, each client is more concerned about the clustering results on its own data.

Inspired by the above, we propose a novel federated deep subspace clustering model, i.e., FDSC, which trains an auto-encoder to learn data features and a self-expressive layer to construct affinity matrix for spectral clustering. Auto-encoder of each client consists of a shared encoder and a private decoder. The shared encoder is communicated between clients and the central server to improve the data feature extraction. The local decoder is used to reconstruct the original data. In order to keep the local feature space structures of affinity matrix, we add adjacency graph information to the network. We align the affinity matrix with the local adjacency matrix in the form of regular terms. Our main contributions are as follows.

\begin{enumerate}
    \item We propose a novel federated deep subspace clustering framework, i.e., FDSC, show in Fig. \ref{fig1}. We extract data features through the shared convolutional encoder, and perform spectral clustering through the affinity matrix obtained from the local self-expressive layer. 
    \item Our studies achieve the maintenance of the graph structure between local data points that the affinity matrix should be aligned with the adjacency matrix of data points.
    \item We evaluate the clustering result of FDSC on four image datasets. Our experiments show that FDSC has better performance than the state-of-the-art methods.
\end{enumerate}

The rest of the paper is organized as follows. In Section 2, we introduce the previous research on deep subspace clustering and federated learning. We show the proposed framework FDSC in Section 3. In Section 4, we present the evaluation results on four public image datasets and compare them with the state-of-the-art methods. Section 5 concludes the framework of this paper.

\section{Related Work}
In this section, we introduce three related fields in this paper, i.e., deep subspace clustering (DSC), federated learning (FL) and federated clustering (FC).

\subsection{Deep Subspace Clustering}

DSC is used to solve the problem that the performance of subspace clustering usually drops in non-linear data spaces \cite{yang2020residual}. DSC incorporates a self-expressive layer in deep auto-encoder to make the representation in the latent space more suitable for spectral clustering \cite{peng2020deep}. Benefiting from data extraction of convolutional networks (CNN), DSC has been widely studied \cite{abavisani2018deep}. DSC network usually consists of an auto-encoder and a self-expression layer \cite{lei2020deep}. The former is used for learning data representation, and the latter is used for spectral clustering.

Suppose there is a set of unlabeled data points $X=\{x_{i}\}_{i=1,...,n}$, where $x_i \in \mathbb{R}^d$ is the $i$-th data sample and $n$ is the total number of samples. These data points come from $k$ different subspace $\{S_{i}\}_{i=1,...,k}$. In general, the self-expressive layer can be expressed as that the data point in a subspace is a linear combination of other points in the same subspace \cite{haeffele2020critique}, i.e. $X=XC$, where $C \in \mathbb{R}^{n \times n}$ is the self-expressive matrix with diagonal block structure \cite{ji2014efficient}. We define the task of the self-expressive layer as
\begin{equation}
\min_{C \in \mathbb{R}^{n \times n}} \Vert C \Vert_p + \frac{\lambda}{2} \Vert X-XC \Vert_F^2 \quad s.t. \quad ( {\rm diag} (C)=0),
\label{eq1}
\end{equation}
where $\lambda$ is a parameter, $\Vert C \Vert_p$ represents an arbitrary norm of matrix C and $\Vert \cdot \Vert_F$ represents the Fibonacci norm. $p$ has been defined in some studies, e.g., \cite{elhamifar2013sparse} studied the sparse subspace clustering with $l_1$ norm, \cite{chen2020stochastic} proposed a stochastic sparse subspace clustering method using $l_2$ norms, \cite{chen2021generalized} performed the generalized nonconvex low-rank tensor approximation with nuclear norm. The self-expressive layer takes the representation of the data extracted by the deep encoder as input. Suppose $\widehat{X}$ is the data reconstructed by the auto-encoder, and $Z$ is the data representation of the encoder output. In general, the loss of deep subspace clustering is \cite{ji2017deep}
\begin{equation}
\begin{aligned}
    &\frac{1}{2} \Vert X-\widehat{X} \Vert_F^2 + \lambda_1 \Vert C \Vert_p + \frac{\lambda_2}{2} \Vert Z-ZC \Vert_F^2 \quad \\
    &s.t. \quad ( {\rm diag} (C)=0),
\end{aligned}
\label{eq2}
\end{equation}
where $\lambda_1$ and $\lambda_2$ are two parameters.

\subsection{Federated Learning}

FL is a decentralized distributed machine learning framework for joint cooperation and privacy protection \cite{nguyen2021federated}. It allows multiple clients with private data to participate in model training through the central server \cite{fang2022robust}. Suppose there are $m$ clients. Client $i$ holds the local datasets with the number $n_i$, i.e. $X^i$. Assume that the local model parameters of client $i$ is $W_i$. Generally, the objective of FL is to learn a global model with collaborative clients \cite{li2019convergence}, 
\begin{equation}
\min_W f(W)\triangleq  \sum_{i=1}^{m}p_{i}f_{i}(W_i),
\label{eq3}
\end{equation}
where $p_{j} = n_j/\sum_i^m n_i$ denotes the parameter weight of the $j$-th client. $f(W)$ is the loss function of the global model with parameters $W$ and $f_i(W_i)$ is the loss function of $i$-th client. 

Many existing studies are devoted to FL in a supervised environment, such as FedAvg \cite{mcmahan2017communication}, FedProx \cite{li2020federated} and FedProto \cite{tan2022fedproto}. In order to solve the heterogeneity of data among clients, some research has turned to a new perspective recently that divide clients into different groups, i.e., clustered federated learning \cite{mansour2020three}. The author of 
 \cite{ghosh2020efficient} proposed IFCA framework that divides clients into different clusters according to similarity and each cluster learns a sharing model. The author in \cite{long2023multi} developed a multi-center aggregation model that clusters clients according to local model parameters. Although these methods have considered the solution of data heterogeneity mentioned in Section 1, they need experts to label the local data of the clients \cite{lubana2022orchestra}. Actually, the data of terminal equipment is often not marked by experts, which makes supervised learning difficult to train. Federated clustering is the way to solve unlabeled data.

\subsection{Federated Clustering}

Although the research of federated learning in the field of clustering has been studied in the past years, most of them focus on clustering clients. Considering the unsupervised learning environment, the author in \cite{dennis2021heterogeneity} proposed k-FED, which is a one-shot federated clustering framework. They solves the communication cost problem in FL. In addition, the author of \cite{stallmann2022towards} developed a federated fuzzy clustering algorithm to solve the clustering of horizontal data partitions. The research on FC needs more exploration. Inspired by centralized deep subspace clustering, we propose a subspace clustering method in federated environment.

\section{The Proposed Model}
This section mainly introduces our proposed method FDSC, which includes motivation, the objective function, the framework and the optimization algorithm.

\subsection{Motivation}
In general, traditional federated learning aims at learning common data representations from various clients \cite{collins2021exploiting}. This approach enables the representation model to adapt to each private client. The application of deep subspace clustering brings vitality to federated clustering learning. There are three observations in federated deep subspace clustering:

\begin{enumerate}
    \item Traditional federated learning which is to learn a global clustering model does not perform well on the client with data heterogeneity. Personalized federated learning usually needs a shared representation model and a local downstream task model.
    \item The key of deep subspace clustering is to generate the self-expressive matrix for data representation. Since spectral clustering is performed on the self-expressive matrix, the self-expressive layer is local model.
    \item In order to keep the reconstruction of local data, the decoder on client should not be shared.
\end{enumerate}

There is currently a lack of a federated framework for deep subspace learning, which preserves the commonality of data representation on each client and performs subspace clustering locally.


\subsection{Objective Function}
Inspired by the above discoveries, our proposed method FDSC divides the data on the client into different subspaces. Each client has a shared encoder $M_{E}$, a private self-expressive layer $M_{R}$ and a private decoder $M_{D}$. Let $E_i$, $R_i$ and $D_i$ denote the encoder parameters, the self-expressive layer parameters and the decoder parameters of the client $i$. $M_{E_i}$ captures the common features among clients in federated learning framework. $M_{D_i}$ maintains the local feature space of the client. 

Suppose $Z_i$ is the output of the encoder $M_{E_i}$, i.e., the representation matrix of data set $X_i$ on the client $i$. Each $z_i$ in $Z_i$ is a node in the linear space. Through the fully connected linear layer $M_{R_i}$, the weight of self-expressive layer is the self-expressive matrix. The loss function on client $i$ is
\begin{equation}
\begin{aligned}
    &f_i = \frac{1}{2} \Vert X^i-\widehat{X}^i \Vert_F^2 + \lambda_1 \Vert R_i \Vert_p + \frac{\lambda_2}{2} \Vert Z^i-Z^iR_i \Vert_F^2 \\
    &s.t. \quad ( {\rm diag} (R_i)=0),
\end{aligned}
\label{eq4}
\end{equation}
where $\lambda_1$ and $\lambda_2$ are parameters, and $\Vert \cdot \Vert_p$ is arbitrary matrix norm. Then, the objective function of FDSC is defined as
\begin{equation}
\min_{E_*} \sum_{i=1}^{m}p_{i} \min_{E_i,R_i,D_i} f_{i}(E_i,R_i,D_i),
\label{eq5}
\end{equation}
where $p_{j} = n_j/\sum_i^m n_i$ denotes the weight of the $j$-th client. $E_*$ is the global encoder, which is weighted and averaged by the encoders $E_i^t$ of the clients participating in the $t$ round of communication,
\begin{equation}
E_*^{(t+1)} = \sum_{i=1}^{m}p_{i} E_i^t.
\label{eq6}
\end{equation}

In order to ensure that the self-expressive matrix keeps the block characteristics of similar features on the client, we propose an approach of aligning to local adjacency matrix. We define an adjacency matrix constructor, i.e., $g(\cdot)$, and establish an adjacency matrix on local data $X^i$. $g(\cdot)$ can be the $k$-nearest neighbors ($k$-NN) method \cite{peterson2009k}. Then, we have
\begin{equation}
A_i = g_{k-NN}(X^i),
\label{eq7}
\end{equation}
where $A_i$ is the adjacency matrix on the $i$-th client. Then, the objective function on client $i$ is
\begin{equation}
\begin{aligned}
    &f_i = \frac{1}{2} \Vert X^i-\widehat{X}^i \Vert_F^2 + \lambda_1 \Vert R_i \Vert_p \\
    &+ \frac{\lambda_2}{2} \Vert Z^i-Z^iR_i \Vert_F^2 + \lambda_3(\alpha A_i - \beta R_i) \\
&s.t. ( {\rm diag} (R_i)=0),
\end{aligned}
\label{eq8}
\end{equation}
where $\lambda_3$, $\alpha$ and $\beta$ are balance parameters.

\subsection{Framework}

In this study, the auto-encoder is implemented by the convolutional neural networks (CNN) to train image data. To be specific, we use two CNN layers for encoder and decoder respectively and a fully connected layer for self-expressive layer. The overall framework of FDSC is shown in Fig. \ref{fig1}.

\subsection{Optimization Algorithm}

\begin{algorithm}
    \caption{The FDSC Algorithm}
    \label{alg1}
    \textbf{Parameters:} Number of communication rounds $T$, number of local epochs $\tau$, number of clients $m$, participation rate $r$, parameters $\lambda_1, \lambda_2, \lambda_3, \alpha, \beta$. \\
    \textbf{Initialization} $E_*^0, R_1^0, \cdots, R_m^0, D_1^0, \cdots, D_m^0$  \\
    \BlankLine
    \emph{//** Generating adjacency matrix with $K$-NN **//}\\	
    \For{$i=1$ to $m$ in parallel}{
        $A_i = g_{k-NN}(X^i)$ \\
        Store matrix $A_i$ locally  \\
     }
     \emph{//** Server federation **//} \\
     \For{$t=1 $ to $T$}{
        Server chooses a random group of clients $S^t$ of size $rm$ \\
        Send $E_*^{t-1}$ to the clients in $S^t$ \\
        \For{each client $i$ in $S^t$ in parallel}{
            \emph{//** Local training **//}\\
            Client $i$ trains model with $E_*^{t-1}$ \\
            Sever collects result : $E_i^t$ \\
        }
        Server computes the new global encoder by Eq. (\ref{eq6}) \\
     }
     \emph{//** Client training **//} \\
     Client $i$ receives global encoder $E_*^{t-1}$ \\
     Client $i$ initializes $E_i^t=E_*^{t-1}, R_i^t=R_i^{t-1}, D_i^t=D_i^{t-1}$ \\
     \For{$e=1 $ to $\tau$}{
        Train the local model by Eq. (\ref{eq8}) \\ 
     }
     Client $i$ sends $E_i^t$ to the server \\
\end{algorithm}

In FDSC, the sever is responsible for communication among clients. The server is used to calculate and broadcast the global encoder $E_*$. On each client, the local model consists of shared encoder $E_i$, private self-expression layer $R_i$ and private decoder $D_i$. At the beginning of local training, the encoder $E_i$ is replaced by the received global encoder $E_*$. To compute $\lambda_3(\alpha A_i - \beta R_i)$ in Eq. (\ref{eq8}), we use $K$-NN to calculate the adjacency matrix of each client data and store it before model training. In order to ensure the stability of local model training, we use stochastic gradient descent with momentum. The details of FDSC are given in Algorithm 1.

\section{Experiment}
 This section shows the experimental results of the proposed FDSC on four public image datasets.

\begin{figure*}[t]
\centering
\includegraphics[width=1.0\textwidth]{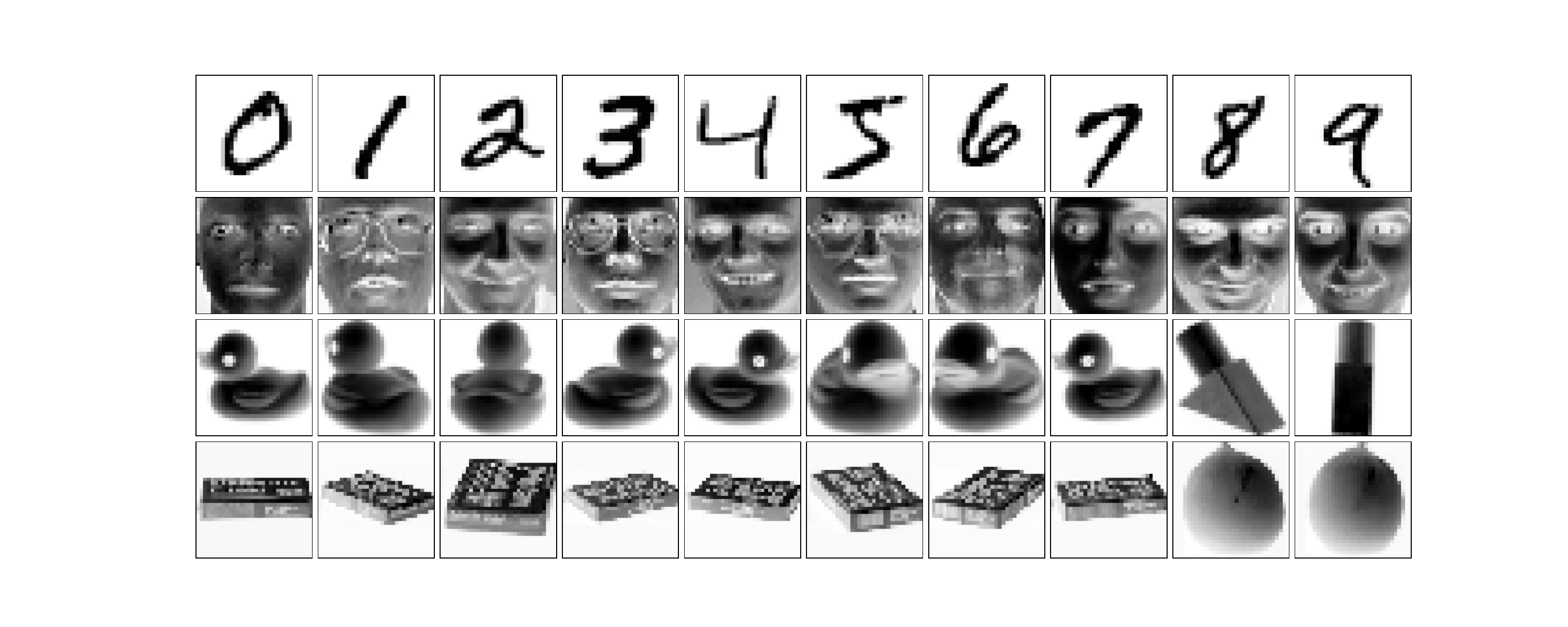}
\caption{The samples of image datasets MNIST, ORL, COIL20 and COIL100.}
\label{fig:dataset}
\end{figure*}

\subsection{The Used Datasets}
In order to compare with the state-of-the-art subspace clustering methods, we perform FDSC on four image datasets. The datasets are shown as follows.

\textbf{MNIST\footnote{http://yann.lecun.com/exdb/mnist/}}. This dataset consists of black and white images of handwritten digits $0-9$. The whole dataset has 70,000 images, of which the training set contains 60,000 images and the test set has 10,000 images. Each image has $28\times28$ gray pixels. In this clustering task, we only use the training set. The first row in Fig.\ref{fig:dataset} shows several example images.

\textbf{ORL\footnote{https://www.cs.columbia.edu/CAVE/software/softlib/coil-100.php}}. This dataset consists of face images, including 40 different samples. And there are 10 images in each sample directory, which are collected in different time, illumination, facial expressions and facial details. Each image has the size of $112\times92$ pixels. We compress the image size to $32\times32$ pixels. The last row in Fig.\ref{fig:dataset} shows some images.

\textbf{COIL20\footnote{https://www.cs.columbia.edu/CAVE/software/softlib/coil-20.php}}. This dataset is a collection of color pictures, which includes 20 objects from different angles, taking an image every 5 degrees, and each object has 72 images. The size of each image is $128\times128$ pixels. Some images are shown in the second row in Fig.\ref{fig:dataset}.  

\textbf{COIL100\footnote{https://www.cs.columbia.edu/CAVE/software/softlib/coil-100.php}}. This dataset is composed of three-channel color images of 100 objects. Each object has 72 postures, which are taken at different angles in a 360 rotation. Each image has the size of $128\times128$ pixels. The third row in Fig.\ref{fig:dataset} illustrates example images.

\subsection{Experiment Settings}
In order to evaluate the clustering performance of our method, we compared FDSC with Low rank subspace clustering 
 (LRSC) \cite{vidal2014low}, Deep Low-Rank Subspace Clustering (DLRSC) \cite{kheirandishfard2020deep} and Deep Subspace Clustering Networks (DSCN) \cite{ji2017deep}. We divided the dataset containing $n$ samples into $m$ subsets, where each subset contains about $n/m$ samples. We randomly selected samples with $q$ categories into these subsets, where $1 \leq q \leq c$ and $c$ is the total number of classes in the data set. In our experiment, we set $\tau=7, \lambda_3 = 1e6, \alpha=1, \beta=1$. The other setting of training parameters are as follows: for MNIST, $T=100, \lambda_1 = 1, \lambda_2=15, m=20, r=0.25$; for ORL, $T=200, \lambda_1 = 2, \lambda_2=0.2, m=5, r=0.4$; for COIL20, $T=100, \lambda_1 = 1, \lambda_2=75, m=5, r=0.4$; for COIL100, $T=100, \lambda_1 = 1, \lambda_2=15, m=5, r=0.4$.

In order to compare the effect of clustering, we used four evaluation indicators, which are ACC, NMI, AMI and ARI. We compared the clustering labels with the real labels to calculate the clustering accuracy, i.e.,
\begin{equation}
ACC = \frac{1}{n}\sum_{i=1}^m \sum_{j=1}^{n_i} \delta(o(\hat{y}_j),y_j)
\label{eq9}
\end{equation}
where $y_j$ is the real label of sample $x_j$ and $\hat{y}_j$ is the clustering label in experiments. Indicator function $\delta(x,y)=1$ when $x=y$, otherwise $0$. $o(\hat{y}_j)$ is a mapping function to find the clustering label that best match the true label.

NMI is normalised mutual information, which is a commonly used evaluation metric in clustering, it measures the degree of consistency between the clustering results and the real labels, the NMI index value ranges from [0, 1], the closer the value is to 1, the higher the degree of similarity between the clustering results and the real labels, the NMI value is 0 means that there is no correlation between the clustering results and the real labels, i.e., clustering effect is very bad, and NMI value is 1 means that the clustering results are completely consistent with the real labels. NMI of 1 means that the clustering results are completely consistent with the real labels.

The AMI index is an adapted form of the Mutual Information Indicator (MII), which corrects for the uncertainty of the clustering results by taking into account the effects of random assignment and category imbalance on the results. Like the NMI, the AMI index takes values in the range [0, 1], with values closer to 1 indicating better clustering results.

ARI is the Adjusted Rand Index, which is used to measure the similarity between the clustering results and the real labels, and it takes into account the effect of the random allocation of data, and takes the value in the range of [-1, 1].The closer the value of ARI index is to 1, it means that the clustering results are more similar to the real labels. 

The final results of the four clustering indexes in this section are presented in the form of percentage.
\subsection{Representation Visualization}
\begin{figure*}[t]
\centering
\includegraphics[width=1\textwidth]{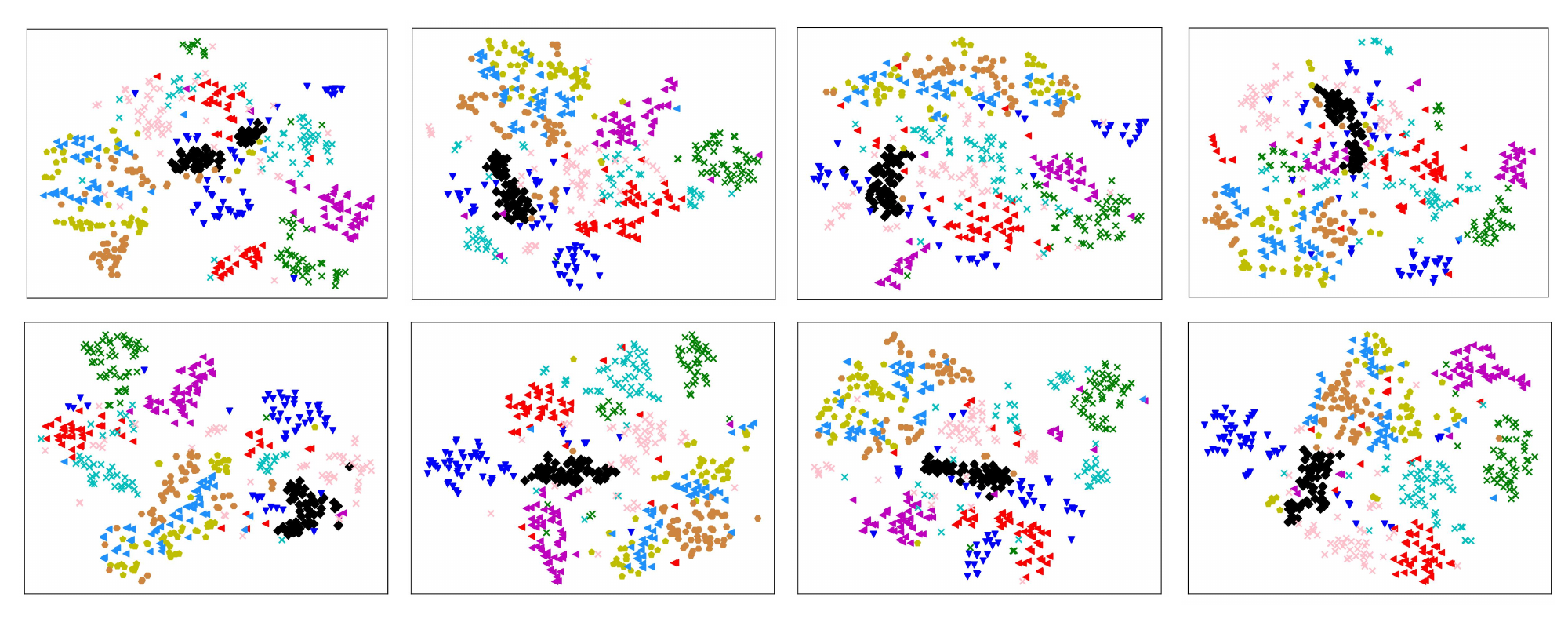}
\caption{Scatterplot of 2D representation of DSCN at the first row and FDSC at the second row. Each column represents a separate representation of the same data on the same client. Images of different colors represent different categories of data.}
\label{fig:representation}
\end{figure*}

This subsection aims to explore the influence of federated subspace clustering. We used MNIST to train FDSC and DSCN to display clustering results of data representation. In FDSC, we used $20$ clients to train image samples, and $q=10$. Then, we used t-SNE method \cite{van2008visualizing} to reduce the representation of the image to 2D. Fig.\ref{fig:representation}  shows the 2D scatter plots of $4$ clients randomly selected from all $20$ participants.The first row shows the result of FDSC representation, and the second row shows the result of DSCN representation, where the data points with different colours represent the real categories that the data belongs to.

On the first client, aggregation of two types of data points, royal blue and brown, using FSDC, works better. On the second client, aggregation of data points using FSDC's royal blue, teal, and brown colors works better. On the third client, pink data point aggregation with FSDC works better. On the fourth client, the aggregation of two types of data points, teal and brown, works better with FSDC.As shown in Fig.\ref{fig:representation}, the represented scatter plot of FDSC has a better subspace segmentation effect than the centralized DSCN.

In addition, we evaluated the clustering results of this 2D scatter points with ACC, NMI, AMI and ARI. The first row in Table \ref{table:repre} shows that the representation results on four clients with using the local DSCN method. The second row in Table \ref{table:repre} shows the representation scatter points on FDSC. Table \ref{table:repre} shows four measurements of 2D points of FDSC and DSCN.
As is shown, under the four clustering evaluation indexes, the corresponding values of FDSC were greater than those of DSCN.The clustering effect of 2D representation points on FDSC is better than that on DSCN. Our proposed FDSC absorbs the features between clients, which enhances the clustering effect.

\begin{table}[htbp]
	\centering
        \label{tab:tsne}
	\caption{Cluster evaluations of 2D points. The larger the data, the better the clustering effect.}
	\begin{tabular}{ccccc}
		\toprule  
		Methods&ACC&NMI&AMI&ARI \\ 
		\cmidrule(r){2-5}
		DSCN&39.1&43.3&40.7&22.5 \\
            FDSC&63.4&59.2&57.6&45.4 \\
		\bottomrule  
	\end{tabular}
        \label{table:repre}
\end{table}

\subsection{Self-expression Matrix Visualization}

\begin{figure}[t]
 \begin{minipage}{0.48\linewidth}
     \vspace{3pt}  
     \centerline{\includegraphics[width=\textwidth]{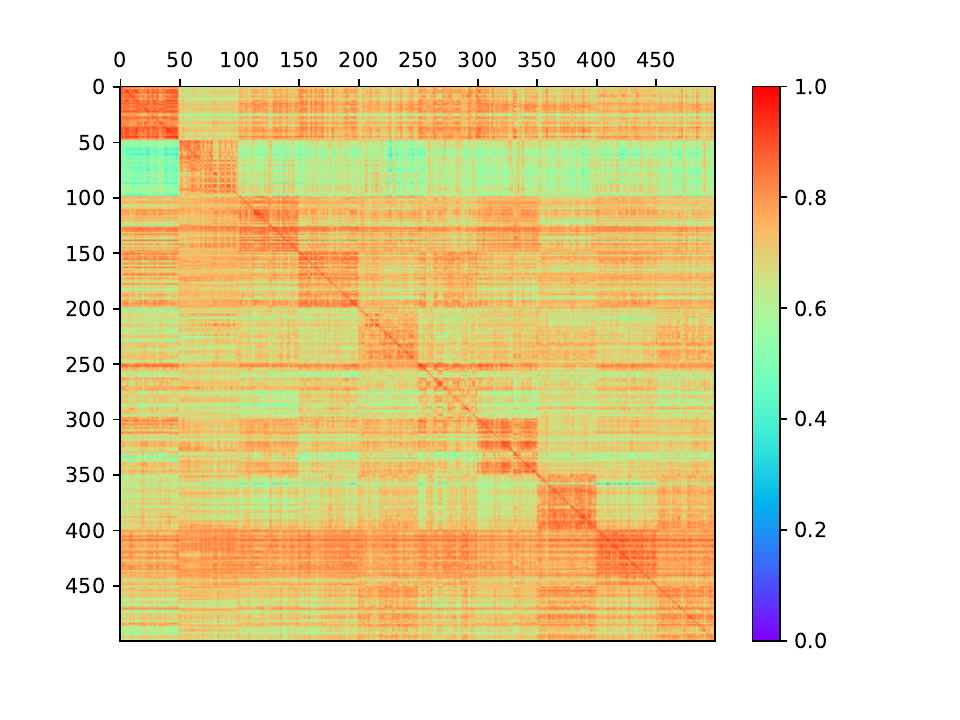}}
     \vspace{3pt}
     \centerline{\includegraphics[width=\textwidth]{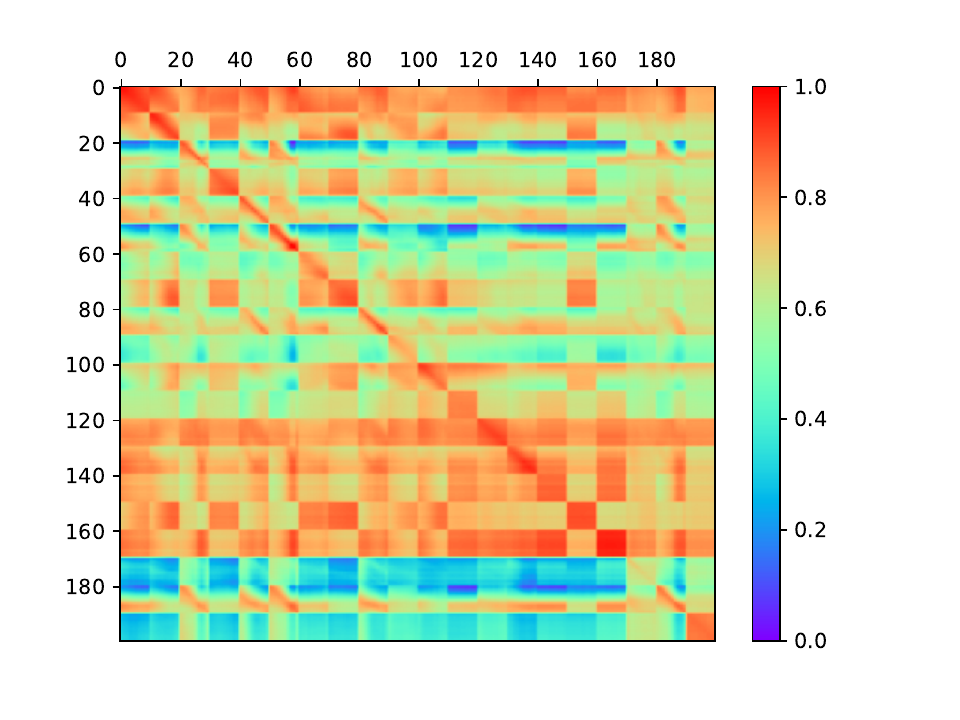}}
     \vspace{3pt}
 \end{minipage}
   \begin{minipage}{0.48\linewidth}
     \vspace{3pt}  
     \centerline{\includegraphics[width=\textwidth]{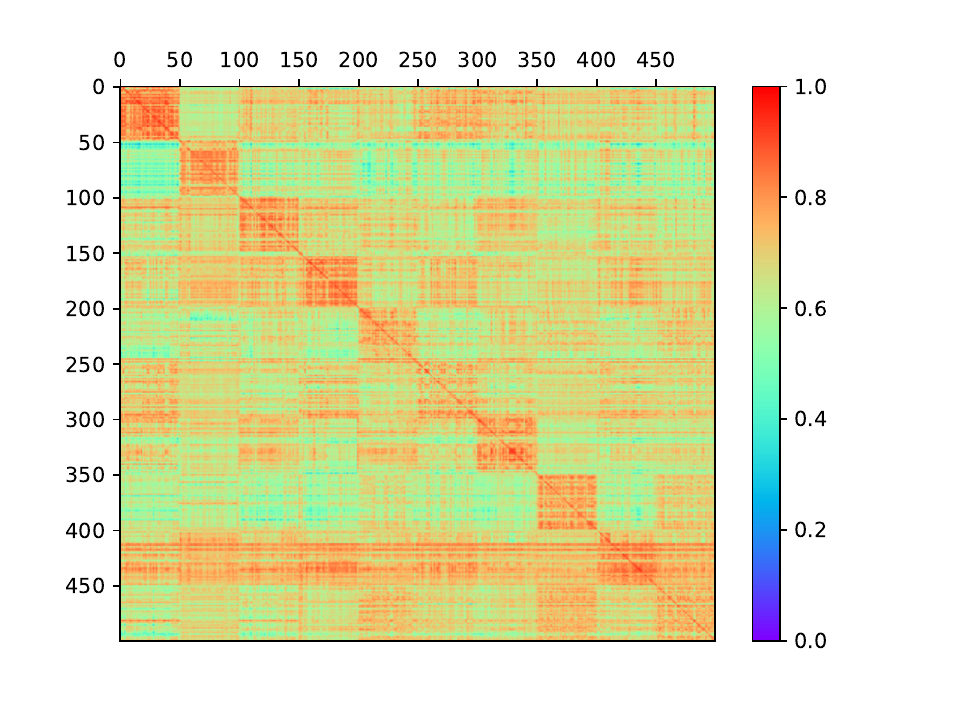}}
     \vspace{3pt}
     \centerline{\includegraphics[width=\textwidth]{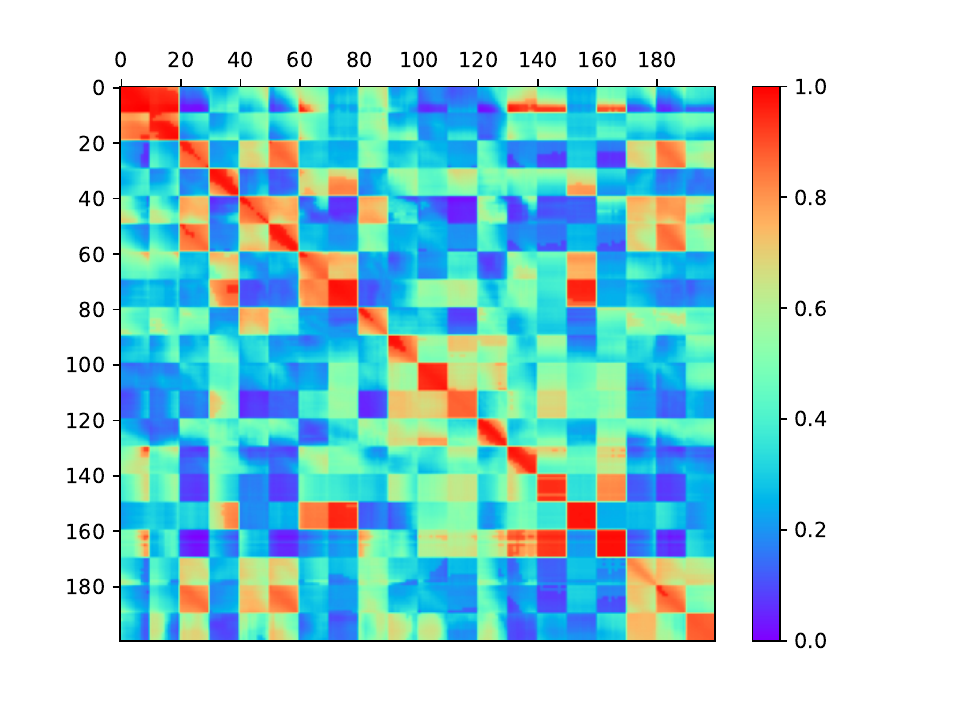}}
     \vspace{3pt}
 \end{minipage}
	\caption{Visualization of self-expression matrices of DSCN at the first column and FDSC at the second column. The first row is the training result on MNIST and the second row is the training result on COIL20. Colors indicate the data in the different classes.}
	\label{fig:matrix}
\end{figure}

This section explores the influence of federated self-expression learning with FDSC on clustering. We used MNIST and COIL20 to train FDSC and visualized the weight matrix of the self-expression layer.We set up 20 participants,the number of classes for each client is q = 10 for MNIST and q = 20 for COIL20. For comparison, the same client data is also trained on the DSCN method. Fig. \ref{fig:matrix} illustrates the results of self-expression matrix with local DSCN and FDSC respectively. The first row shows the matrix training on MNIST and the seconda row shows the results on COIL20.The matrix on the left of each row is the self-expression matrix of the DSCN, and the matrix results of the FDSC are on the right. 

When using the MNIST dataset, we distribute 10,000 test samples to 20 participants, so that each participant gets exactly 50 samples for each category. When using the COIL20 dataset, we assign 20 classes of data to each participant, and 10 data points for each class of data to facilitate testing. Subsequently, on the two types of datasets, we averaged the clustering results of the same category of data among the 20 participants and expressed the average results as numerical values between 0 and 1.

The color legend on the right side of each figure in Fig. \ref{fig:matrix} shows the relationship between color and value. The more red the color patch is, the stronger the similarity of the corresponding subspace data. The more the color patch tends to be purple, the weaker the data similarity of the corresponding subspace. In the first row, the diagonal patch of the second image is closer to red than the other area patches. In contrast, the color difference between diagonal and non-diagonal patches in the first image is not significant. In the second row, the diagonal patches of the second image are very close to red in color, while the non-diagonal patches are predominantly blue and green. In the first image, the diagonal patches are not as close to red as in the second image, but the non-diagonal patches are relatively close to red. This indicates that the self-expression matrix trained by FDSC has a better block diagonal structure, that is, the federated deep subspace learning method can better represent a certain data as a linear combination of other data in the same subspace.

\subsection{Clustering Results}


\begin{table*} 
  \centering
  \caption{Results of image clustering by using all mentioned methods. By setting different $q$, we test ACC, NMI, AMI and ARI on four datasets respectively. The data set used is in the first row, the value of $q$ is in the second row, the evaluation index is in the third row, and the clustering results of different methods (100 for perfect aggregation) are in 4-8 rows.}
  \label{tab:evaluation}
  \begin{tabular}{ccccccccccccccccc}
  \hline
            \multirow{3}{*}{Methods}&\multicolumn{8}{c}{MNIST}& \multicolumn{8}{c}{ORL} \\ 
            \cmidrule(lr){2-9} \cmidrule(lr){10-17} 
            
            \multirow{3}{*}{ }&\multicolumn{4}{c}{q=5}& \multicolumn{4}{c}{q=10} &\multicolumn{4}{c}{q=20}&\multicolumn{4}{c}{q=40}  \\
            \cmidrule(lr){2-5} \cmidrule(lr){6-9} \cmidrule(lr){10-13} \cmidrule(lr){14-17}

            \multirow{3}{*}{ }& ACC & NMI & AMI & ARI & ACC & NMI & AMI & ARI & ACC & NMI & AMI & ARI & ACC & NMI & AMI & ARI  \\ 

            \hline 
            \textbf{LRSC} & 77.4 & 78.6 & 79.1 & 68.2 & 67.9 & 72.1 & 71.6 & 65.8 &50.8&66.9 &34.5 &28.5 &62.1 & 73.5 & 38.9 &34.7\\
            \textbf{DLRSC} & 79.6 & 80.2 & 80.9 & 69.3 & 68.5 & 74.3 & 73.7& 65.2 &51.4 &67.3 &34.1 &27.9 &62.6 &75.4 &38.6 &34.2\\
            \textbf{DSCN}&81.7 & 82.3 & 81.3 &71.3& 70.4 & 77.5 & 76.1 &66.3 & 52.5 & 69.2& 36.6 & 28.7 &64.5 &76.1&40.1 &35.2 \\ 
            \hline 
            \textbf{FDSC1} & 82.4 & 84.1& 83.6 & 72.8 & 71.7 & 78.1 & 77.2 &63.1 &53.2 &70.7 &36.4 &28.5 &65.2 &76.5 & 41.6 &36.1 \\
            \textbf{FDSC2} & 83.8 & 84.6 & 84.4 & 73.1 & 72.9 & 79.7 & 78.9& 65.8 &53.7 &71.6 &37.4 &29.3 &67.5 &77.6 &42.4 &37.4\\
            \hline
  \end{tabular}
\end{table*}

\begin{table*}
  \centering
  \begin{tabular}{ccccccccccccccccc}
  \hline
            \multirow{3}{*}{Methods}&\multicolumn{8}{c}{COIL20}& \multicolumn{8}{c}{COIL100} \\ 
            \cmidrule(lr){2-9} \cmidrule(lr){10-17} 
            
            \multirow{3}{*}{ }&\multicolumn{4}{c}{q=10}& \multicolumn{4}{c}{q=20} &\multicolumn{4}{c}{q=50}&\multicolumn{4}{c}{q=100}  \\
            \cmidrule(lr){2-5} \cmidrule(lr){6-9} \cmidrule(lr){10-13} \cmidrule(lr){14-17}

            \multirow{3}{*}{ }& ACC & NMI & AMI & ARI & ACC & NMI & AMI & ARI & ACC & NMI & AMI & ARI & ACC & NMI & AMI & ARI  \\ 

            \hline 
            \textbf{LRSC} &64.2 &72.8 &69.3 &50.4 &68.7 & 81.3 &75.2 & 58.4 &60.7 &80.3&71.3&54.8 &75.3&86.4 &72.6&58.4\\
            \textbf{DLRSC} &66.3 &73.9 &70.1 &51.8 &68.2 &81.9 &76.4 &60.8 &62.9 &81.5 &72.6&55.4 &76.4&87.9&73.5&60.2\\
            \textbf{DSCN}& 68.5 &76.7 &71.2 &53.5 &70.5 &84.2 &78.5 &61.4 &64.7 &83.4 &73.9 &56.5 &78.1 &89.4 &75.2 &61.6 \\ 
            \hline 
            \textbf{FDSC1} &70.5 &73.9 &73.1 &54.7 &71.3 &85.4 &79.8 &62.5 &65.8 &84.3 & 75.3 &58.9&78.4 &90.5 &76.8 &63.4 \\
            \textbf{FDSC2} & 71.3 &75.2 &74.1 &55.3 &72.5 &86.6 &80.6 &63.6 &66.7 &85.7 &76.4 &59.5 &79.4 &91.5 &77.2 &64.9\\
            \hline
  \end{tabular}
\end{table*}

Table \ref{tab:evaluation} shows image clustering results on four datasets by using LRSC, DLRSC, DSCN, FDSC1($\lambda_3=0$) and FDSC2($\lambda_3=1e6$). We assumed that the number of clients corresponding to each dataset is $m$ and the data categories held on each client is $q$. To ensure the appropriateness of the amount of data obtained by each client when using different datasets,we set $m=20$ on MNIST, $m=5$ on ORL, $m=5$ on COIL20 and $m=5$ on COIL100.To explore the influence of data heterogeneity on training effect, we first tested clustering methods with different $q$ to compare the performance of the methods. We set $q=\{5,10\}$ on MNIST, $q=\{20,40\}$ on ORL, $q=\{10,20\}$ on COIL20 and $q=\{50,100\}$ on COIL100.

As is shown, our federated method FDSC has higher clustering accuracy than other methods with centralized learning on all datasets. Under various parameter settings, FDSC achieves the best clustering index. Compared with FDSC1, FDSC2 adds a regular term that aligns the self-expression matrix to the adjacency matrix, and the clustering result is better, which verifies the FDSC with regular optimization achieves better clustering results than the method without regular optimization. In the environment of data heterogeneity, in addition to MNIST data sets, other data sets such as ORL, COIL20 and COIL100 perform better when the degree of data heterogeneity on the client is low, that is, the $q$ is large. This shows that in most cases, with the reduction of client data heterogeneity, FDSC method can effectively improve the clustering effect. On MNIST, when the data heterogeneity of the client is higher, i.e., the $q$ is less, the clustering result is better. 


\subsection{Parameter Discussion}

\begin{table}[h]
\centering
\caption{The clustering results by different number of clients. The higher the value corresponding to the clustering index, the better the clustering effect.}
\label{tab:numberofclients}
\begin{tabular}{cccccc}
\toprule
    Number of clients&Methods&ACC&NMI &AMI   &ARI\\ 
    \midrule
    \multirow{2}*{10} & FDSC1 & 86.9 &81.5&81.2&79.3 \\
        & FDSC2  & 87.6	&82.7&82.5&79.8\\  
    \hline
    \multirow{2}*{20}  & FDSC1 & 87.5 &83.2&83.1 &78.6\\
        & FDSC2  & 88.4	&84.3&84.2&79.8\\  
    \hline
    \multirow{2}*{30} & FDSC1 &88.2&85.1&84.7&82.1\\
        & FDSC2 &90.6&86.5&86.9&82.8\\
    \hline
    \multirow{2}*{40} & FDSC1&90.3&86.7&87.9&83.5\\
        & FDSC2&91.9&87.3&88.1&84.2\\
    \hline
    \multirow{2}*{50} & FDSC1&92.6&87.9&87.4&83.6\\
        & FDSC2&93.5&89.1&89.5&85.3\\
    \bottomrule
\end{tabular}
\end{table}

In this subsection, we made more insights into the impacts of the parameters in FDSC, i.e., the number of clients $m$ and the batch size $b$. We used MNIST to explore the influence of parameters on FDSC.

To probe the influence of the number of clients, we set the batch size of clients to 500, and changed the number of clients $m$ with $\{10,20,30,40,50\}$. Table \ref{tab:numberofclients} shows the clustering results by different $m$. From the results, as the number of clients increases, the clustering effect is better.The clustering accuracy of FDSC benefits from the participation of more clients. 

\begin{table}
\centering
\caption{The clustering results by different batch size. The higher the value corresponding to the clustering index, the better the clustering effect.}
\label{tab:numberofbatch}
\begin{tabular}{cccccc}
\toprule
    Batch size&Methods&ACC&NMI &AMI   &ARI\\ 
    \midrule
    \multirow{2}*{100} & FDSC1 & 82.5 &80.6&80.9&74.7 \\
        & FDSC2  & 83.2	&81.5&81.6&75.2\\  
    \hline
    \multirow{2}*{250}  & FDSC1 & 84.2 &82.7&82.4 &76.8\\
        & FDSC2  & 85.8	&83.4&82.9&77.3\\  
    \hline
    \multirow{2}*{500} & FDSC1 &86.1&83.9&83.2&77.6\\
        & FDSC2 &87.4&84.1&83.8&78.2\\
    \hline
    \multirow{2}*{750} & FDSC1&87.8&83.6&83.3&79.2\\
        & FDSC2&89.2&85.0&85.6&80.3\\
    \hline
    \multirow{2}*{1000} & FDSC1&91.9&87.4&87.9&81.3\\
        & FDSC2&93.2&89.9&89.7&82.6\\
    \bottomrule
\end{tabular}
\end{table}

To examine the effect of the batch size in model training, we assigned the MNIST to 20 clients by varying batchsize $b$ with $\{100,250,500,750,1000\}$. Table \ref{tab:numberofbatch} shows the evaluation indicators in FDSC. As is shown, as the batch size increases, the clustering effect is better.FDSC achieves higher performance with the increase of client batch size.

\section{Discussion and Conclusion}
This study proposes a novel federated deep subspace clustering framework, called FDSC. As far as we know, FDSC is the first federated clustering model that using deep subspace clustering to group the data on all clients jointly. FDSC exploits shared encoder to learn common representation and learns the self-expressive matrix for local spectral clustering task. We finally evaluated FDSC on four image data sets and compared it with subspace clustering model. The experimental results show that FDSC performs better under the used indexes. In the future, big models like the vision transformer \cite{liu2021swin} will be considered to enhance the global model in the server.


\section{Acknowledgments}
This work was supported in part by the National Natural Science Foundation of China (Nos. 62272392 and U22A2025), the Key Research and Development Program of Shaanxi Province (No. 2023-YBGY-405), and Fundamental Research Funds for the Central Universities (No. G2023KY0603).





\bibliographystyle{ACM-Reference-Format}
\bibliography{refs}


\begin{thebibliography}{45}


\ifx \showCODEN    \undefined \def \showCODEN     #1{\unskip}     \fi
\ifx \showDOI      \undefined \def \showDOI       #1{#1}\fi
\ifx \showISBNx    \undefined \def \showISBNx     #1{\unskip}     \fi
\ifx \showISBNxiii \undefined \def \showISBNxiii  #1{\unskip}     \fi
\ifx \showISSN     \undefined \def \showISSN      #1{\unskip}     \fi
\ifx \showLCCN     \undefined \def \showLCCN      #1{\unskip}     \fi
\ifx \shownote     \undefined \def \shownote      #1{#1}          \fi
\ifx \showarticletitle \undefined \def \showarticletitle #1{#1}   \fi
\ifx \showURL      \undefined \def \showURL       {\relax}        \fi
\providecommand\bibfield[2]{#2}
\providecommand\bibinfo[2]{#2}
\providecommand\natexlab[1]{#1}
\providecommand\showeprint[2][]{arXiv:#2}

\bibitem[Abavisani and Patel(2018)]%
        {abavisani2018deep}
\bibfield{author}{\bibinfo{person}{Mahdi Abavisani} {and}
  \bibinfo{person}{Vishal~M Patel}.} \bibinfo{year}{2018}\natexlab{}.
\newblock \showarticletitle{Deep multimodal subspace clustering networks}.
\newblock \bibinfo{journal}{\emph{IEEE Journal of Selected Topics in Signal
  Processing}} \bibinfo{volume}{12}, \bibinfo{number}{6}
  (\bibinfo{year}{2018}), \bibinfo{pages}{1601--1614}.
\newblock


\bibitem[Arikumar et~al\mbox{.}(2022)]%
        {arikumar2022fl}
\bibfield{author}{\bibinfo{person}{KS Arikumar}, \bibinfo{person}{Sahaya~Beni
  Prathiba}, \bibinfo{person}{Mamoun Alazab}, \bibinfo{person}{Thippa~Reddy
  Gadekallu}, \bibinfo{person}{Sharnil Pandya}, \bibinfo{person}{Javed~Masood
  Khan}, {and} \bibinfo{person}{Rajalakshmi~Shenbaga Moorthy}.}
  \bibinfo{year}{2022}\natexlab{}.
\newblock \showarticletitle{FL-PMI: federated learning-based person movement
  identification through wearable devices in smart healthcare systems}.
\newblock \bibinfo{journal}{\emph{Sensors}} \bibinfo{volume}{22},
  \bibinfo{number}{4} (\bibinfo{year}{2022}), \bibinfo{pages}{1377}.
\newblock


\bibitem[Baek et~al\mbox{.}(2021)]%
        {baek2021deep}
\bibfield{author}{\bibinfo{person}{Sangwon Baek}, \bibinfo{person}{Gangjoon
  Yoon}, \bibinfo{person}{Jinjoo Song}, {and} \bibinfo{person}{Sang~Min Yoon}.}
  \bibinfo{year}{2021}\natexlab{}.
\newblock \showarticletitle{Deep self-representative subspace clustering
  network}.
\newblock \bibinfo{journal}{\emph{Pattern Recognition}}  \bibinfo{volume}{118}
  (\bibinfo{year}{2021}), \bibinfo{pages}{108041}.
\newblock


\bibitem[Banabilah et~al\mbox{.}(2022)]%
        {banabilah2022federated}
\bibfield{author}{\bibinfo{person}{Syreen Banabilah}, \bibinfo{person}{Moayad
  Aloqaily}, \bibinfo{person}{Eitaa Alsayed}, \bibinfo{person}{Nida Malik},
  {and} \bibinfo{person}{Yaser Jararweh}.} \bibinfo{year}{2022}\natexlab{}.
\newblock \showarticletitle{Federated learning review: Fundamentals, enabling
  technologies, and future applications}.
\newblock \bibinfo{journal}{\emph{Information processing \& management}}
  \bibinfo{volume}{59}, \bibinfo{number}{6} (\bibinfo{year}{2022}),
  \bibinfo{pages}{103061}.
\newblock


\bibitem[Berahmand et~al\mbox{.}(2022)]%
        {berahmand2022novel}
\bibfield{author}{\bibinfo{person}{Kamal Berahmand}, \bibinfo{person}{Mehrnoush
  Mohammadi}, \bibinfo{person}{Azadeh Faroughi}, {and}
  \bibinfo{person}{Rojiar~Pir Mohammadiani}.} \bibinfo{year}{2022}\natexlab{}.
\newblock \showarticletitle{A novel method of spectral clustering in attributed
  networks by constructing parameter-free affinity matrix}.
\newblock \bibinfo{journal}{\emph{Cluster Computing}} (\bibinfo{year}{2022}),
  \bibinfo{pages}{1--20}.
\newblock


\bibitem[Chen et~al\mbox{.}(2020)]%
        {chen2020stochastic}
\bibfield{author}{\bibinfo{person}{Ying Chen}, \bibinfo{person}{Chun-Guang Li},
  {and} \bibinfo{person}{Chong You}.} \bibinfo{year}{2020}\natexlab{}.
\newblock \showarticletitle{Stochastic sparse subspace clustering}. In
  \bibinfo{booktitle}{\emph{Proceedings of the IEEE/CVF conference on computer
  vision and pattern recognition}}. \bibinfo{pages}{4155--4164}.
\newblock


\bibitem[Chen et~al\mbox{.}(2021)]%
        {chen2021generalized}
\bibfield{author}{\bibinfo{person}{Yongyong Chen}, \bibinfo{person}{Shuqin
  Wang}, \bibinfo{person}{Chong Peng}, \bibinfo{person}{Zhongyun Hua}, {and}
  \bibinfo{person}{Yicong Zhou}.} \bibinfo{year}{2021}\natexlab{}.
\newblock \showarticletitle{Generalized nonconvex low-rank tensor approximation
  for multi-view subspace clustering}.
\newblock \bibinfo{journal}{\emph{IEEE Transactions on Image Processing}}
  \bibinfo{volume}{30} (\bibinfo{year}{2021}), \bibinfo{pages}{4022--4035}.
\newblock


\bibitem[Collins et~al\mbox{.}(2021)]%
        {collins2021exploiting}
\bibfield{author}{\bibinfo{person}{Liam Collins}, \bibinfo{person}{Hamed
  Hassani}, \bibinfo{person}{Aryan Mokhtari}, {and} \bibinfo{person}{Sanjay
  Shakkottai}.} \bibinfo{year}{2021}\natexlab{}.
\newblock \showarticletitle{Exploiting shared representations for personalized
  federated learning}. In \bibinfo{booktitle}{\emph{International Conference on
  Machine Learning}}. PMLR, \bibinfo{pages}{2089--2099}.
\newblock


\bibitem[Dennis et~al\mbox{.}(2021)]%
        {dennis2021heterogeneity}
\bibfield{author}{\bibinfo{person}{Don~Kurian Dennis}, \bibinfo{person}{Tian
  Li}, {and} \bibinfo{person}{Virginia Smith}.}
  \bibinfo{year}{2021}\natexlab{}.
\newblock \showarticletitle{Heterogeneity for the win: One-shot federated
  clustering}. In \bibinfo{booktitle}{\emph{International Conference on Machine
  Learning}}. PMLR, \bibinfo{pages}{2611--2620}.
\newblock


\bibitem[Elhamifar and Vidal(2013)]%
        {elhamifar2013sparse}
\bibfield{author}{\bibinfo{person}{Ehsan Elhamifar} {and}
  \bibinfo{person}{Ren{\'e} Vidal}.} \bibinfo{year}{2013}\natexlab{}.
\newblock \showarticletitle{Sparse subspace clustering: Algorithm, theory, and
  applications}.
\newblock \bibinfo{journal}{\emph{IEEE transactions on pattern analysis and
  machine intelligence}} \bibinfo{volume}{35}, \bibinfo{number}{11}
  (\bibinfo{year}{2013}), \bibinfo{pages}{2765--2781}.
\newblock


\bibitem[Fang and Ye(2022)]%
        {fang2022robust}
\bibfield{author}{\bibinfo{person}{Xiuwen Fang} {and} \bibinfo{person}{Mang
  Ye}.} \bibinfo{year}{2022}\natexlab{}.
\newblock \showarticletitle{Robust federated learning with noisy and
  heterogeneous clients}. In \bibinfo{booktitle}{\emph{Proceedings of the
  IEEE/CVF Conference on Computer Vision and Pattern Recognition}}.
  \bibinfo{pages}{10072--10081}.
\newblock


\bibitem[Gao et~al\mbox{.}(2020)]%
        {gao2020privacy}
\bibfield{author}{\bibinfo{person}{Dashan Gao}, \bibinfo{person}{Ben Tan},
  \bibinfo{person}{Ce Ju}, \bibinfo{person}{Vincent~W Zheng}, {and}
  \bibinfo{person}{Qiang Yang}.} \bibinfo{year}{2020}\natexlab{}.
\newblock \showarticletitle{Privacy threats against federated matrix
  factorization}.
\newblock \bibinfo{journal}{\emph{arXiv preprint arXiv:2007.01587}}
  (\bibinfo{year}{2020}).
\newblock


\bibitem[Ghosh et~al\mbox{.}(2020)]%
        {ghosh2020efficient}
\bibfield{author}{\bibinfo{person}{Avishek Ghosh}, \bibinfo{person}{Jichan
  Chung}, \bibinfo{person}{Dong Yin}, {and} \bibinfo{person}{Kannan
  Ramchandran}.} \bibinfo{year}{2020}\natexlab{}.
\newblock \showarticletitle{An efficient framework for clustered federated
  learning}.
\newblock \bibinfo{journal}{\emph{Advances in Neural Information Processing
  Systems}}  \bibinfo{volume}{33} (\bibinfo{year}{2020}),
  \bibinfo{pages}{19586--19597}.
\newblock


\bibitem[Haeffele et~al\mbox{.}(2020)]%
        {haeffele2020critique}
\bibfield{author}{\bibinfo{person}{Benjamin~D Haeffele}, \bibinfo{person}{Chong
  You}, {and} \bibinfo{person}{Ren{\'e} Vidal}.}
  \bibinfo{year}{2020}\natexlab{}.
\newblock \showarticletitle{A critique of self-expressive deep subspace
  clustering}.
\newblock \bibinfo{journal}{\emph{arXiv preprint arXiv:2010.03697}}
  (\bibinfo{year}{2020}).
\newblock


\bibitem[Han et~al\mbox{.}(2022)]%
        {han2022fedx}
\bibfield{author}{\bibinfo{person}{Sungwon Han}, \bibinfo{person}{Sungwon
  Park}, \bibinfo{person}{Fangzhao Wu}, \bibinfo{person}{Sundong Kim},
  \bibinfo{person}{Chuhan Wu}, \bibinfo{person}{Xing Xie}, {and}
  \bibinfo{person}{Meeyoung Cha}.} \bibinfo{year}{2022}\natexlab{}.
\newblock \showarticletitle{FedX: Unsupervised Federated Learning with Cross
  Knowledge Distillation}. In \bibinfo{booktitle}{\emph{Computer Vision--ECCV
  2022: 17th European Conference, Tel Aviv, Israel, October 23--27, 2022,
  Proceedings, Part XXX}}. Springer, \bibinfo{pages}{691--707}.
\newblock


\bibitem[Ji et~al\mbox{.}(2014)]%
        {ji2014efficient}
\bibfield{author}{\bibinfo{person}{Pan Ji}, \bibinfo{person}{Mathieu Salzmann},
  {and} \bibinfo{person}{Hongdong Li}.} \bibinfo{year}{2014}\natexlab{}.
\newblock \showarticletitle{Efficient dense subspace clustering}. In
  \bibinfo{booktitle}{\emph{IEEE Winter conference on applications of computer
  vision}}. IEEE, \bibinfo{pages}{461--468}.
\newblock


\bibitem[Ji et~al\mbox{.}(2017)]%
        {ji2017deep}
\bibfield{author}{\bibinfo{person}{Pan Ji}, \bibinfo{person}{Tong Zhang},
  \bibinfo{person}{Hongdong Li}, \bibinfo{person}{Mathieu Salzmann}, {and}
  \bibinfo{person}{Ian Reid}.} \bibinfo{year}{2017}\natexlab{}.
\newblock \showarticletitle{Deep subspace clustering networks}.
\newblock \bibinfo{journal}{\emph{Advances in neural information processing
  systems}}  \bibinfo{volume}{30} (\bibinfo{year}{2017}).
\newblock


\bibitem[Kheirandishfard et~al\mbox{.}(2020)]%
        {kheirandishfard2020deep}
\bibfield{author}{\bibinfo{person}{Mohsen Kheirandishfard},
  \bibinfo{person}{Fariba Zohrizadeh}, {and} \bibinfo{person}{Farhad
  Kamangar}.} \bibinfo{year}{2020}\natexlab{}.
\newblock \showarticletitle{Deep low-rank subspace clustering}. In
  \bibinfo{booktitle}{\emph{Proceedings of the IEEE/CVF conference on computer
  vision and pattern recognition workshops}}. \bibinfo{pages}{864--865}.
\newblock


\bibitem[Lei et~al\mbox{.}(2020)]%
        {lei2020deep}
\bibfield{author}{\bibinfo{person}{Jianjun Lei}, \bibinfo{person}{Xinyu Li},
  \bibinfo{person}{Bo Peng}, \bibinfo{person}{Leyuan Fang},
  \bibinfo{person}{Nam Ling}, {and} \bibinfo{person}{Qingming Huang}.}
  \bibinfo{year}{2020}\natexlab{}.
\newblock \showarticletitle{Deep spatial-spectral subspace clustering for
  hyperspectral image}.
\newblock \bibinfo{journal}{\emph{IEEE Transactions on Circuits and Systems for
  Video Technology}} \bibinfo{volume}{31}, \bibinfo{number}{7}
  (\bibinfo{year}{2020}), \bibinfo{pages}{2686--2697}.
\newblock


\bibitem[Li et~al\mbox{.}(2021)]%
        {li2021self}
\bibfield{author}{\bibinfo{person}{Kun Li}, \bibinfo{person}{Yao Qin},
  \bibinfo{person}{Qiang Ling}, \bibinfo{person}{Yingqian Wang},
  \bibinfo{person}{Zaiping Lin}, {and} \bibinfo{person}{Wei An}.}
  \bibinfo{year}{2021}\natexlab{}.
\newblock \showarticletitle{Self-supervised deep subspace clustering for
  hyperspectral images with adaptive self-expressive coefficient matrix
  initialization}.
\newblock \bibinfo{journal}{\emph{IEEE Journal of Selected Topics in Applied
  Earth Observations and Remote Sensing}}  \bibinfo{volume}{14}
  (\bibinfo{year}{2021}), \bibinfo{pages}{3215--3227}.
\newblock


\bibitem[Li et~al\mbox{.}(2020)]%
        {li2020federated}
\bibfield{author}{\bibinfo{person}{Tian Li}, \bibinfo{person}{Anit~Kumar Sahu},
  \bibinfo{person}{Manzil Zaheer}, \bibinfo{person}{Maziar Sanjabi},
  \bibinfo{person}{Ameet Talwalkar}, {and} \bibinfo{person}{Virginia Smith}.}
  \bibinfo{year}{2020}\natexlab{}.
\newblock \showarticletitle{Federated optimization in heterogeneous networks}.
\newblock \bibinfo{journal}{\emph{Proceedings of Machine learning and systems}}
   \bibinfo{volume}{2} (\bibinfo{year}{2020}), \bibinfo{pages}{429--450}.
\newblock


\bibitem[Li et~al\mbox{.}(2019)]%
        {li2019convergence}
\bibfield{author}{\bibinfo{person}{Xiang Li}, \bibinfo{person}{Kaixuan Huang},
  \bibinfo{person}{Wenhao Yang}, \bibinfo{person}{Shusen Wang}, {and}
  \bibinfo{person}{Zhihua Zhang}.} \bibinfo{year}{2019}\natexlab{}.
\newblock \showarticletitle{On the convergence of fedavg on non-iid data}.
\newblock \bibinfo{journal}{\emph{arXiv preprint arXiv:1907.02189}}
  (\bibinfo{year}{2019}).
\newblock


\bibitem[Li et~al\mbox{.}(2022)]%
        {li2022neural}
\bibfield{author}{\bibinfo{person}{Zengyi Li}, \bibinfo{person}{Yubei Chen},
  \bibinfo{person}{Yann LeCun}, {and} \bibinfo{person}{Friedrich~T Sommer}.}
  \bibinfo{year}{2022}\natexlab{}.
\newblock \showarticletitle{Neural manifold clustering and embedding}.
\newblock \bibinfo{journal}{\emph{arXiv preprint arXiv:2201.10000}}
  (\bibinfo{year}{2022}).
\newblock


\bibitem[Liu et~al\mbox{.}(2021)]%
        {liu2021swin}
\bibfield{author}{\bibinfo{person}{Ze Liu}, \bibinfo{person}{Yutong Lin},
  \bibinfo{person}{Yue Cao}, \bibinfo{person}{Han Hu}, \bibinfo{person}{Yixuan
  Wei}, \bibinfo{person}{Zheng Zhang}, \bibinfo{person}{Stephen Lin}, {and}
  \bibinfo{person}{Baining Guo}.} \bibinfo{year}{2021}\natexlab{}.
\newblock \showarticletitle{Swin transformer: Hierarchical vision transformer
  using shifted windows}. In \bibinfo{booktitle}{\emph{Proceedings of the
  IEEE/CVF international conference on computer vision}}.
  \bibinfo{pages}{10012--10022}.
\newblock


\bibitem[Long et~al\mbox{.}(2023)]%
        {long2023multi}
\bibfield{author}{\bibinfo{person}{Guodong Long}, \bibinfo{person}{Ming Xie},
  \bibinfo{person}{Tao Shen}, \bibinfo{person}{Tianyi Zhou},
  \bibinfo{person}{Xianzhi Wang}, {and} \bibinfo{person}{Jing Jiang}.}
  \bibinfo{year}{2023}\natexlab{}.
\newblock \showarticletitle{Multi-center federated learning: clients clustering
  for better personalization}.
\newblock \bibinfo{journal}{\emph{World Wide Web}} \bibinfo{volume}{26},
  \bibinfo{number}{1} (\bibinfo{year}{2023}), \bibinfo{pages}{481--500}.
\newblock


\bibitem[Lubana et~al\mbox{.}(2022)]%
        {lubana2022orchestra}
\bibfield{author}{\bibinfo{person}{Ekdeep~Singh Lubana},
  \bibinfo{person}{Chi~Ian Tang}, \bibinfo{person}{Fahim Kawsar},
  \bibinfo{person}{Robert~P Dick}, {and} \bibinfo{person}{Akhil Mathur}.}
  \bibinfo{year}{2022}\natexlab{}.
\newblock \showarticletitle{Orchestra: Unsupervised federated learning via
  globally consistent clustering}.
\newblock \bibinfo{journal}{\emph{arXiv preprint arXiv:2205.11506}}
  (\bibinfo{year}{2022}).
\newblock


\bibitem[Lv et~al\mbox{.}(2021)]%
        {lv2021pseudo}
\bibfield{author}{\bibinfo{person}{Juncheng Lv}, \bibinfo{person}{Zhao Kang},
  \bibinfo{person}{Xiao Lu}, {and} \bibinfo{person}{Zenglin Xu}.}
  \bibinfo{year}{2021}\natexlab{}.
\newblock \showarticletitle{Pseudo-supervised deep subspace clustering}.
\newblock \bibinfo{journal}{\emph{IEEE Transactions on Image Processing}}
  \bibinfo{volume}{30} (\bibinfo{year}{2021}), \bibinfo{pages}{5252--5263}.
\newblock


\bibitem[Mansour et~al\mbox{.}(2020)]%
        {mansour2020three}
\bibfield{author}{\bibinfo{person}{Yishay Mansour}, \bibinfo{person}{Mehryar
  Mohri}, \bibinfo{person}{Jae Ro}, {and} \bibinfo{person}{Ananda~Theertha
  Suresh}.} \bibinfo{year}{2020}\natexlab{}.
\newblock \showarticletitle{Three approaches for personalization with
  applications to federated learning}.
\newblock \bibinfo{journal}{\emph{arXiv preprint arXiv:2002.10619}}
  (\bibinfo{year}{2020}).
\newblock


\bibitem[McMahan et~al\mbox{.}(2017)]%
        {mcmahan2017communication}
\bibfield{author}{\bibinfo{person}{Brendan McMahan}, \bibinfo{person}{Eider
  Moore}, \bibinfo{person}{Daniel Ramage}, \bibinfo{person}{Seth Hampson},
  {and} \bibinfo{person}{Blaise~Aguera y Arcas}.}
  \bibinfo{year}{2017}\natexlab{}.
\newblock \showarticletitle{Communication-efficient learning of deep networks
  from decentralized data}. In \bibinfo{booktitle}{\emph{The Proceedings of
  Artificial intelligence and statistics}}. PMLR, \bibinfo{pages}{1273--1282}.
\newblock


\bibitem[Ng et~al\mbox{.}(2001)]%
        {ng2001spectral}
\bibfield{author}{\bibinfo{person}{Andrew Ng}, \bibinfo{person}{Michael
  Jordan}, {and} \bibinfo{person}{Yair Weiss}.}
  \bibinfo{year}{2001}\natexlab{}.
\newblock \showarticletitle{On spectral clustering: Analysis and an algorithm}.
\newblock \bibinfo{journal}{\emph{Advances in neural information processing
  systems}}  \bibinfo{volume}{14} (\bibinfo{year}{2001}).
\newblock


\bibitem[Nguyen et~al\mbox{.}(2021)]%
        {nguyen2021federated}
\bibfield{author}{\bibinfo{person}{Dinh~C Nguyen}, \bibinfo{person}{Ming Ding},
  \bibinfo{person}{Pubudu~N Pathirana}, \bibinfo{person}{Aruna Seneviratne},
  \bibinfo{person}{Jun Li}, {and} \bibinfo{person}{H~Vincent Poor}.}
  \bibinfo{year}{2021}\natexlab{}.
\newblock \showarticletitle{Federated learning for internet of things: A
  comprehensive survey}.
\newblock \bibinfo{journal}{\emph{IEEE Communications Surveys \& Tutorials}}
  \bibinfo{volume}{23}, \bibinfo{number}{3} (\bibinfo{year}{2021}),
  \bibinfo{pages}{1622--1658}.
\newblock


\bibitem[Nguyen et~al\mbox{.}(2022)]%
        {nguyen2022federated}
\bibfield{author}{\bibinfo{person}{Dinh~C Nguyen}, \bibinfo{person}{Quoc-Viet
  Pham}, \bibinfo{person}{Pubudu~N Pathirana}, \bibinfo{person}{Ming Ding},
  \bibinfo{person}{Aruna Seneviratne}, \bibinfo{person}{Zihuai Lin},
  \bibinfo{person}{Octavia Dobre}, {and} \bibinfo{person}{Won-Joo Hwang}.}
  \bibinfo{year}{2022}\natexlab{}.
\newblock \showarticletitle{Federated learning for smart healthcare: A survey}.
\newblock \bibinfo{journal}{\emph{ACM Computing Surveys (CSUR)}}
  \bibinfo{volume}{55}, \bibinfo{number}{3} (\bibinfo{year}{2022}),
  \bibinfo{pages}{1--37}.
\newblock


\bibitem[Peng et~al\mbox{.}(2020)]%
        {peng2020deep}
\bibfield{author}{\bibinfo{person}{Xi Peng}, \bibinfo{person}{Jiashi Feng},
  \bibinfo{person}{Joey~Tianyi Zhou}, \bibinfo{person}{Yingjie Lei}, {and}
  \bibinfo{person}{Shuicheng Yan}.} \bibinfo{year}{2020}\natexlab{}.
\newblock \showarticletitle{Deep subspace clustering}.
\newblock \bibinfo{journal}{\emph{IEEE transactions on neural networks and
  learning systems}} \bibinfo{volume}{31}, \bibinfo{number}{12}
  (\bibinfo{year}{2020}), \bibinfo{pages}{5509--5521}.
\newblock


\bibitem[Peterson(2009)]%
        {peterson2009k}
\bibfield{author}{\bibinfo{person}{Leif~E Peterson}.}
  \bibinfo{year}{2009}\natexlab{}.
\newblock \showarticletitle{K-nearest neighbor}.
\newblock \bibinfo{journal}{\emph{Scholarpedia}} \bibinfo{volume}{4},
  \bibinfo{number}{2} (\bibinfo{year}{2009}), \bibinfo{pages}{1883}.
\newblock


\bibitem[Sattler et~al\mbox{.}(2020)]%
        {sattler2020clustered}
\bibfield{author}{\bibinfo{person}{Felix Sattler},
  \bibinfo{person}{Klaus-Robert M{\"u}ller}, {and} \bibinfo{person}{Wojciech
  Samek}.} \bibinfo{year}{2020}\natexlab{}.
\newblock \showarticletitle{Clustered federated learning: Model-agnostic
  distributed multitask optimization under privacy constraints}.
\newblock \bibinfo{journal}{\emph{IEEE transactions on neural networks and
  learning systems}} \bibinfo{volume}{32}, \bibinfo{number}{8}
  (\bibinfo{year}{2020}), \bibinfo{pages}{3710--3722}.
\newblock


\bibitem[Stallmann and Wilbik(2022)]%
        {stallmann2022towards}
\bibfield{author}{\bibinfo{person}{Morris Stallmann} {and}
  \bibinfo{person}{Anna Wilbik}.} \bibinfo{year}{2022}\natexlab{}.
\newblock \showarticletitle{Towards Federated Clustering: A Federated Fuzzy $ c
  $-Means Algorithm (FFCM)}.
\newblock \bibinfo{journal}{\emph{arXiv preprint arXiv:2201.07316}}
  (\bibinfo{year}{2022}).
\newblock


\bibitem[Tan et~al\mbox{.}(2022)]%
        {tan2022fedproto}
\bibfield{author}{\bibinfo{person}{Yue Tan}, \bibinfo{person}{Guodong Long},
  \bibinfo{person}{Lu Liu}, \bibinfo{person}{Tianyi Zhou},
  \bibinfo{person}{Qinghua Lu}, \bibinfo{person}{Jing Jiang}, {and}
  \bibinfo{person}{Chengqi Zhang}.} \bibinfo{year}{2022}\natexlab{}.
\newblock \showarticletitle{Fedproto: Federated prototype learning across
  heterogeneous clients}. In \bibinfo{booktitle}{\emph{Proceedings of the AAAI
  Conference on Artificial Intelligence}}, Vol.~\bibinfo{volume}{36}.
  \bibinfo{pages}{8432--8440}.
\newblock


\bibitem[Van~der Maaten and Hinton(2008)]%
        {van2008visualizing}
\bibfield{author}{\bibinfo{person}{Laurens Van~der Maaten} {and}
  \bibinfo{person}{Geoffrey Hinton}.} \bibinfo{year}{2008}\natexlab{}.
\newblock \showarticletitle{Visualizing data using t-SNE.}
\newblock \bibinfo{journal}{\emph{Journal of machine learning research}}
  \bibinfo{volume}{9}, \bibinfo{number}{11} (\bibinfo{year}{2008}).
\newblock


\bibitem[Vidal and Favaro(2014)]%
        {vidal2014low}
\bibfield{author}{\bibinfo{person}{Ren{\'e} Vidal} {and} \bibinfo{person}{Paolo
  Favaro}.} \bibinfo{year}{2014}\natexlab{}.
\newblock \showarticletitle{Low rank subspace clustering (LRSC)}.
\newblock \bibinfo{journal}{\emph{Pattern Recognition Letters}}
  \bibinfo{volume}{43} (\bibinfo{year}{2014}), \bibinfo{pages}{47--61}.
\newblock


\bibitem[Yang et~al\mbox{.}(2019)]%
        {yang2019federated}
\bibfield{author}{\bibinfo{person}{Qiang Yang}, \bibinfo{person}{Yang Liu},
  \bibinfo{person}{Tianjian Chen}, {and} \bibinfo{person}{Yongxin Tong}.}
  \bibinfo{year}{2019}\natexlab{}.
\newblock \showarticletitle{Federated machine learning: Concept and
  applications}.
\newblock \bibinfo{journal}{\emph{ACM Transactions on Intelligent Systems and
  Technology (TIST)}} \bibinfo{volume}{10}, \bibinfo{number}{2}
  (\bibinfo{year}{2019}), \bibinfo{pages}{1--19}.
\newblock


\bibitem[Yang et~al\mbox{.}(2020)]%
        {yang2020residual}
\bibfield{author}{\bibinfo{person}{Shuai Yang}, \bibinfo{person}{Wenqi Zhu},
  {and} \bibinfo{person}{Yuesheng Zhu}.} \bibinfo{year}{2020}\natexlab{}.
\newblock \showarticletitle{Residual encoder-decoder network for deep subspace
  clustering}. In \bibinfo{booktitle}{\emph{2020 IEEE International Conference
  on Image Processing (ICIP)}}. IEEE, \bibinfo{pages}{2895--2899}.
\newblock


\bibitem[Zhang et~al\mbox{.}(2016)]%
        {zhang2016spectral}
\bibfield{author}{\bibinfo{person}{Hongyan Zhang}, \bibinfo{person}{Han Zhai},
  \bibinfo{person}{Liangpei Zhang}, {and} \bibinfo{person}{Pingxiang Li}.}
  \bibinfo{year}{2016}\natexlab{}.
\newblock \showarticletitle{Spectral--spatial sparse subspace clustering for
  hyperspectral remote sensing images}.
\newblock \bibinfo{journal}{\emph{IEEE Transactions on Geoscience and Remote
  Sensing}} \bibinfo{volume}{54}, \bibinfo{number}{6} (\bibinfo{year}{2016}),
  \bibinfo{pages}{3672--3684}.
\newblock


\bibitem[Zhang et~al\mbox{.}(2018)]%
        {zhang2018survey}
\bibfield{author}{\bibinfo{person}{Qingchen Zhang}, \bibinfo{person}{Laurence~T
  Yang}, \bibinfo{person}{Zhikui Chen}, {and} \bibinfo{person}{Peng Li}.}
  \bibinfo{year}{2018}\natexlab{}.
\newblock \showarticletitle{A survey on deep learning for big data}.
\newblock \bibinfo{journal}{\emph{Information Fusion}}  \bibinfo{volume}{42}
  (\bibinfo{year}{2018}), \bibinfo{pages}{146--157}.
\newblock


\bibitem[Zhang et~al\mbox{.}(2017)]%
        {zhang2017low}
\bibfield{author}{\bibinfo{person}{Yupei Zhang}, \bibinfo{person}{Ming Xiang},
  {and} \bibinfo{person}{Bo Yang}.} \bibinfo{year}{2017}\natexlab{}.
\newblock \showarticletitle{Low-rank preserving embedding}.
\newblock \bibinfo{journal}{\emph{Pattern Recognition}}  \bibinfo{volume}{70}
  (\bibinfo{year}{2017}), \bibinfo{pages}{112--125}.
\newblock


\bibitem[Zhu et~al\mbox{.}(2019)]%
        {zhu2019multi}
\bibfield{author}{\bibinfo{person}{Pengfei Zhu}, \bibinfo{person}{Binyuan Hui},
  \bibinfo{person}{Changqing Zhang}, \bibinfo{person}{Dawei Du},
  \bibinfo{person}{Longyin Wen}, {and} \bibinfo{person}{Qinghua Hu}.}
  \bibinfo{year}{2019}\natexlab{}.
\newblock \showarticletitle{Multi-view deep subspace clustering networks}.
\newblock \bibinfo{journal}{\emph{arXiv preprint arXiv:1908.01978}}
  (\bibinfo{year}{2019}).
\newblock


\end{thebibliography}

\appendix





\end{document}